# Spatio-Temporal Avoidance of Predicted Occupancy in Human-Robot Collaboration

Jared Flowers, Marco Faroni, Gloria Wiens, Nicola Pedrocchi

*Abstract*— This paper addresses human-robot collaboration (HRC) challenges of integrating predictions of human activity to provide a proactive-n-reactive response capability for the robot. Prior works that consider current or predicted human poses as static obstacles are too nearsighted or too conservative in planning, potentially causing delayed robot paths. Alternatively, time-varying prediction of human poses would enable robot paths that avoid anticipated human poses, synchronized dynamically in time and space. Herein, a proactive path planning method, denoted STAP, is presented that uses spatio-temporal human occupancy maps to find robot trajectories that anticipate human movements, allowing robot passage without stopping. In addition, STAP anticipates delays from robot speed restrictions required by ISO/TS 15066 speed and separation monitoring (SSM). STAP also proposes a sampling-based planning algorithm based on RRT* to solve the spatio-temporal motion planning problem and find paths of minimum expected duration. Experimental results show STAP generates paths of shorter duration and greater average robot-human separation distance throughout tasks. Additionally, STAP more accurately estimates robot trajectory durations in HRC, which are useful in arriving at proactive-n-reactive robot sequencing.

## I. INTRODUCTION

A prevalent challenge in human-robot collaboration (HRC) is providing robot(s) proactive-n-reactive capability to correctly anticipate and respond to humans working in close proximity. In contexts such as manufacturing, robot(s) and human(s) typically perform cyclic tasks in which human motion is likely repetitive and predictable. If HRC controller algorithms, including motion planners, can efficiently leverage such knowledge, then the robot can proactively avoid anticipated interruptions. Additionally, the robot could take advantage of predicted windows of time that are free of obstruction to move safely and seamlessly among humans. A motion planner that utilizes the predicted motions of humans can also estimate a 'robotic task completion time' metric useful in selecting the next robot task to perform. The planner could then further modify the timing of robot motions to reduce the amount of time a robot spends in close proximity to a human, allowing greater human comfort.

*Funding was provided by the NSF/NRI: INT: COLLAB: Manufacturing USA: Intelligent Human-Robot Collaboration for Smart Factory (Award I.D. #:1830383).

J. Flowers and G. Wiens are with Mechanical & Aerospace Engineering, University of Florida, Gainesville, FL 32611 USA (e-mail: jared.flowers@ufl.edu, gwiens@ufl.edu).
M. Faroni is with the Department of Robotics of the University of Michigan, Ann Arbor, MI 48109 USA (e-mail: mfaroni@umich.edu)
N. Pedrocchi is with Istituto di Sistemi e Tecnologie Industriali Intelligenti per il Manifatturiero Avanzato Consiglio Nazionale delle Ricerche, Milan, Italy (e-mail:, Nicola.pedrocchi@stiima.cnr.it).

State-of-the-art variations of Rapidly Exploring Random Trees (RRT) and Probabilistic Roadmaps have included features that allow reaction to an obstacle's current pose while computing fast enough for online planning [1-5]. These recent variations only consider current, static poses of dynamic obstacles. Recent human-aware planners consider static, anticipated human volumes and robot speed reduction due to human proximity to generate paths specifically beneficial to HRC [6-8]. Human-aware planners also consider dynamic obstacles as having anticipated, static poses. Since these planners only consider static obstacle poses and do not consider obstacles' time varying motion, they are nearsighted in planning. This nearsightedness allows obstacles to drive the robot toward temporary entrapments and production delay. For example, consider a workcell with two assembly stations, a robot that services both stations, and a human that alternates between stations. If the robot's path planner considers human pose as static, then the human and robot could interfere with each other as the human moves between stations. The method herein permits robot paths that proactively account for the anticipated human motion between stations, allowing uninterrupted robot and human motion.

This paper proposes the Spatio-Temporal Avoidance of Predictions (STAP) method. The goal of STAP is to reduce the interference between humans and robots during collaborative tasks. STAP includes predictions of the humans' movements in a motion planning problem and searches for a trajectory that avoids potential collisions. While previous methods have also used time-based cost functions, STAP has key differences from previous works. First, STAP uses predicted human motion in 3D space to estimate time-varying human occupancy in a robotic workcell. Second, STAP formulates robot trajectory planning as an optimal motion planning problem in which trajectory execution time under the spatio-temporal avoidance constraints is minimized. A variation of RRT* is developed to solve this planning problem. Third, STAP incorporates the speed and separation monitoring (SSM) rules from the ISO/TS 15066 standard to estimate the effect of predicted human poses on robot speed and trajectory duration [9]. Experiments show that trajectories planned with STAP have significantly shorter execution times and larger robot/human separation distances during close collaboration.

The following sections include a discussion of the STAP method's key steps, experimental validation, results, and conclusions. Section III includes subsections: III.A) detailing the predicted human model, III.B) avoidance intervals, III.C) incorporation of idle times owed to safety stops and slowdowns, III.D) variation of RRT* to utilize avoidance intervals, and III.E) STAP time-parameterization.

## II. RELATED WORKS

Prior planning methods incorporate predicted obstacle motion to proactively avoid anticipated obstructions. One option is to dynamically modify the robot's trajectory based on immediate human motion [10,11]. Another method generates collision free waypoints when time is included in the configuration space and uses a local planner to ensure reachability [12]. Other works define safe intervals for passage through configuration nodes to avoid obstacles when generating 2D robot paths, in which both obstacles and robot are defined as points [13,14]. One planning framework models probabilistic human motion sequences using time sequences of Gaussian Mixture Models (GMMs) [15,16]. It uses the STOMP path optimizer to deform straight-line robot paths to minimize penetration into the human's volume. Other methods use optimization frameworks to generate robot trajectories that maintain a distance between the robot's trajectory and learned time-sequences of human arm motion [17,18]. Some approaches avoid anticipated points of collision for the entire task by considering previously occupied space or predicted human occupancy volumes (HOVs) as static obstacles throughout the task [6,19]. STAP differs from prior works by utilizing predicted, dynamic motions of humans over an entire task rather than static definitions or instantaneous estimates.

In an HRC manufacturing workcell, the structure of tasks facilitates prediction of human motion. The human predictions are inputs to the method presented in this paper. One motion prediction method uses GMMs to generate time sequences of probabilistic models for human arm extension trajectories [20,21]. Then it uses the GMMs for classification and extrapolation for current human motion. GMMs were also used to learn a linear dynamic model to represent human trajectories [22,23]. Probabilistic principal component analysis has also been used to learn motion models for human trajectories, detect motion onset, estimate human speed, and select a motion model to infer a future trajectory [24]. Methods such as Risk of Passage and the Swept Volumes method anticipate volumes at the intersection of robot paths and predicted human motion [25,26].

## III. METHOD

The goal of STAP is to develop a cost function and sampling-based robot path planner variation that considers anticipated, time-varying motion of humans. STAP first generates a spatio-temporal human occupancy map to encode anticipated human poses and times of occupancy. Then STAP considers both the occupancy map and estimated speed reductions enforced by an SSM safety controller to estimate the time required for a robot to traverse robot configurations. Finally, a variation of RRT* is proposed that creates a graph of nodes and estimated path costs using the STAP cost function. STAP outputs the trajectory of minimal duration considering anticipated, time-varying, human occupancy. Fig. 1 shows how STAP fits into an overall robot control scheme as the green blocks. The blue block is the motion prediction method providing input to STAP. Fig. 1 includes the SSM safety controller as the orange block which limits robot speed along the planned trajectory based on real-time robot/human separation distance.

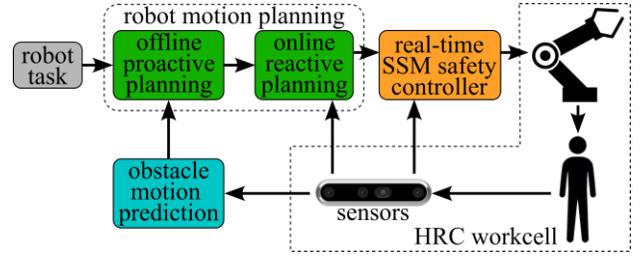

Figure 1. The high-level control loop for an HRC workcell.

### A. Predicted Spatio-Temporal Human-Occupancy Map

First, STAP generates a spatio-temporal occupancy map in cartesian space. Each point of the map stores a set of avoidance intervals, which are the time intervals over which any human body part will occupy that point. The input to STAP is a time series defining human motion in terms of the pelvis location, quaternions of each human body part (aka limb, human skeletal link) w.r.t. the world-z axis, link lengths, and link radii for the human for each time step. At each time step of the human motion, quaternion forward kinematics determines the human joint locations based on the pelvis location, quaternions, and link lengths. Then, the human shape is generalized to a cylinder for each link, shown in fig. 2A. Next, the link cylinders are fit to a discretized 3D grid to generate the blue point cloud in fig. 2B. The cartesian workspace surrounding the robot is denoted by $\mathcal{W}$. Each occupied point in $\mathcal{W}$ is assigned the time of occupancy from the human motion sequence. This results in a set of cartesian points occupied at a list of time steps, denoted $\mathcal{H}$. The set $\mathcal{H}$ can then be used to generate intervals of time over which each point in $\mathcal{W}$ is occupied. Each interval has start time $t_s$ and end time $t_f$. If a point in $\mathcal{H}$ is occupied over a set of consecutive time steps, then the first time step is the start time of an interval and the last time step is the end time. A single point in $\mathcal{H}$ can have multiple intervals, denoted $[t_s, t_f]$, over which the point is occupied.

### B. Avoidance Intervals

The points in $\mathcal{W}$ must inherit occupancy intervals from $\mathcal{H}$. Interval start and end times for a point in $\mathcal{W}$ are denoted $t_{s_{o_i}}$ and $t_{f_{o_i}}$, respectively, for the $i^{th}$ interval resulting from the $o^{th}$ obstacle at that point. The time intervals of occupancy are:

$$\mathcal{A}_{o_i}(x,y,z) = \begin{cases} [t_{s_{o_i}}, t_{f_{o_i}}] & t_{f_o} < t_{end_o} \\ [t_{s_{o_i}}, \infty) & t_{f_o} = t_{end_o} \end{cases}, \quad (1)$$

$$\mathcal{A}(x,y,z) = \cup_o \cup_i \mathcal{A}_{o_i}(x,y,z) \; \forall i \in o, \forall o \in (x,y,z), \quad (2)$$

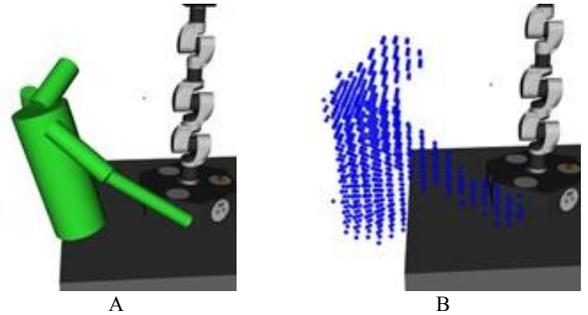

Figure 2. (A) Human pose generalized to cylinders and (B) a point cloud.

where $\mathcal{A}_{o_i}(x,y,z)$ is the $i^{th}$ avoidance interval for the $o^{th}$ obstacle occupying point $(x,y,z)$ and $t_{end_o}$ is the final time step of the predicted motion sequence for the $o^{th}$ obstacle. The $\mathcal{A}(x,y,z)$ is the union of all intervals for which humans occupy point $(x,y,z)$. If point $(x,y,z)$ is occupied by the $o^{th}$ obstacle at the end of the motion prediction for that obstacle, then it must be assumed that point $(x,y,z)$ is occupied by the $o^{th}$ obstacle for the remainder of time. This leads to the notion of a last time of passage for some points in cartesian space:

$$t_{lp_{o_i}}(x,y,z) = \begin{cases} \infty & t_{f_o} < t_{end_o} \\ t_{s_{o_i}} & t_{f_o} = t_{end_o} \end{cases}, \quad (3)$$

$$t_{lp}(x,y,z) = \min_{o \in (x,y,z)} \min_{i \in o} t_{lp_{o_i}}(x,y,z), \quad (4)$$

where $t_{lp_{o_i}}(x,y,z)$ is the last time passage is allowed due to the $i^{th}$ interval for $o^{th}$ obstacle at cartesian point $(x,y,z)$. The $t_{lp}(x,y,z)$ is the last time of passage for point $(x,y,z)$, which is the minimum of $t_{lp_{o_i}}(x,y,z)$ over all intervals for which humans occupy point $(x,y,z)$.

After determining the spatio-temporal occupancy map, a sampling-based planner can generate a trajectory for the robot. The planner generates a new node, having robot configuration $\boldsymbol{q_c}$. Each robot configuration has dimension equal to the degrees of freedom (DOF) of the robot. Connections between the new node and each of its neighbors, having robot configuration $\boldsymbol{q_p}$, are considered. Robot configurations with spacing $\Delta \boldsymbol{q}$ between $\boldsymbol{q_p}$ and $\boldsymbol{q_c}$ must be checked for collision. Let $\boldsymbol{FK}(\boldsymbol{q_p}, \boldsymbol{q_c})$ represent the poses of the robot determined by forward kinematics at configurations between $\boldsymbol{q_p}$ and $\boldsymbol{q_c}$. If an intersection of points in $\mathcal{W}$ and $\boldsymbol{FK}(\boldsymbol{q_p}, \boldsymbol{q_c})$ exists, then the set of avoidance intervals for the intersecting points will be added to the set of avoidance intervals and last pass time for the connection between $\boldsymbol{q_p}$ and $\boldsymbol{q_c}$ in the planner's node graph ($\mathcal{G}$), denoted $\mathcal{A}(\boldsymbol{q_p}, \boldsymbol{q_c})$ and $t_{lp}(\boldsymbol{q_p}, \boldsymbol{q_c})$ respectively:

$$\mathcal{A}(\boldsymbol{q_p}, \boldsymbol{q_c}) = \bigcup_{(x,y,z) \in \boldsymbol{FK}(\boldsymbol{q_p}, \boldsymbol{q_c})} \mathcal{A}(x,y,z), \quad (5)$$

$$t_{lp}(\boldsymbol{q_p}, \boldsymbol{q_c}) = \min_{(x,y,z) \in \boldsymbol{FK}(\boldsymbol{q_p}, \boldsymbol{q_c})} t_{lp}(x,y,z). \quad (6)$$

Next, if $\mathcal{A}(\boldsymbol{q_p}, \boldsymbol{q_c})$ is not an empty set, then the minimum time to reach node $\boldsymbol{q_c}$ from the start configuration while avoiding potential collision between $\boldsymbol{q_p}$ and $\boldsymbol{q_c}$ must be determined. First, the minimum time required to travel between $\boldsymbol{q_p}$ and $\boldsymbol{q_c}$ is:

$$t(\boldsymbol{q_p}, \boldsymbol{q_c}) = \max_{i \in [1,n]} \left| \frac{q_c[i] - q_p[i]}{\bar{q}[i]} \right|, \quad (7)$$

where $\bar{\boldsymbol{q}}$ is the vector of maximum velocities for each robot DOF and $i$ iterates over each DOF $[1, n]$. Each node in $\mathcal{G}$ must store the minimum arrival time for the robot to reach its configuration from the start configuration, denoted $t_{arr_p}$ for node at $\boldsymbol{q_p}$. The $t_{arr}$ for the start node is set to zero. Then the minimum time to reach $\boldsymbol{q_c}$ via $\boldsymbol{q_p}$ is $t_c$:

$$t_c = t_p + t(\boldsymbol{q_p}, \boldsymbol{q_c}), \quad (8)$$

where $t_p$ is initially considered to be $t_{arr_p}$ for the node at $\boldsymbol{q_p}$. This creates a time interval for passage from $\boldsymbol{q_p}$ to $\boldsymbol{q_c}$ of $[t_p, t_c]$. If $[t_p, t_c]$ intersects $\mathcal{A}(\boldsymbol{q_p}, \boldsymbol{q_c})$, then $t_p$ is updated according to:

$$t_p = \min_i \{\mathcal{A}_i(\boldsymbol{q_p}, \boldsymbol{q_c})_{end} + t_{pad} : [t_p, t_c] \cap \mathcal{A}(\boldsymbol{q_p}, \boldsymbol{q_c}) \equiv 0\} \quad (9)$$

The $t_{pad}$ is a user-defined time padding constant. The larger $t_{pad}$ is, the more tolerant the plan is to deviations in actual human motion compared to the prediction. If $t_p$ or $t_c$ is greater than $t_{lp}(\boldsymbol{q_p}, \boldsymbol{q_c})$, then the connection is blocked indefinitely and it is rejected. For example, if the human doesn't back away from the robot after a task, the robot's path could be blocked indefinitely. Every time $t_p$ is updated, $t_c$ must be updated using (8). The $\mathcal{A}(\boldsymbol{q_p}, \boldsymbol{q_c})$ is sorted by increasing interval start time, so the resulting $[t_p, t_c]$ is the earliest time interval when the robot can travel from $\boldsymbol{q_p}$ to $\boldsymbol{q_c}$ without colliding with a predicted human pose. The time to reach $\boldsymbol{q_c}$, denoted $t_c$, can now be considered as the cost to reach $\boldsymbol{q_c}$ for use in finding the optimal path. Now the $t_{arr_c}$ for the node at $\boldsymbol{q_c}$ is also updated to be $t_c$ if $t_c$ is less than $t_{arr_c}$.

*C. Safety-Aware Cost-Function*

When STAP is applied to HRC, the ISO/TS 15066 standard defines modes of robot operation to ensure human safety [9]. The SSM rules of ISO/TS 15066 require that a safety controller, shown as the orange block in fig. 1, enforce a robot speed limit proportional to robot/human separation distance. The safety controller stops the robot if the separation distance becomes too small. To mitigate potential SSM safety controller slowdown effects, STAP adapts its time-based cost function to anticipate speed reductions based on SSM rules applied with the predicted human motion. STAP estimates the effect of SSM by calculating the time between nodes using the formulation from [27]. First, the distance between the $i^{th}$ point on the human at time $t_h$ in the predicted motion ($\boldsymbol{h}_i(t_h)$) and the $j^{th}$ robot point at $\boldsymbol{q}$ (denoted $\boldsymbol{FK}_j(\boldsymbol{q})$) is:

$$D_{ij} = \|\boldsymbol{h}_i(t_h) - \boldsymbol{FK}_j(\boldsymbol{q})\|. \quad (10)$$

The points of the human in $\boldsymbol{h}(t_h)$ are located at the human's joints and centroids of limbs. Then the maximum speed allowed by the $j^{th}$ robot point relative to the $i^{th}$ human point at time $t_h$ and robot configuration $\boldsymbol{q}$ according to SSM is:

$$v_{max}(\boldsymbol{q}, i, j, t_h) =$$

$$\begin{cases} -a_s T_r - v_h + \sqrt{v_h^2 + (a_s T_r)^2 + 2 a_s D_{ij}} & : D_{ij} > \underline{D} \\ 0 & else \end{cases} \quad (11)$$

The $v_h$ is the human's velocity relative to the robot, $a_s$ is the maximum cartesian deceleration of the robot relative to the human, $T_r$ is the reaction time of the robot, and $\underline{D}$ is a minimum distance between human and robot allowed for robot motion. The magnitude of the robot's tangential speed in the direction of the human is:

$$v_{robot}(\boldsymbol{q}, i, j, t_h) = \boldsymbol{J}_j(\boldsymbol{q}) \frac{(\boldsymbol{q_c} - \boldsymbol{q_p})}{t(\boldsymbol{q_p}, \boldsymbol{q_c})} \cdot \frac{(\boldsymbol{h}_i(t_h) - \boldsymbol{FK}_j(\boldsymbol{q}))}{\|\boldsymbol{h}_i(t_h) - \boldsymbol{FK}_j(\boldsymbol{q})\|}, \quad (12)$$

where $\boldsymbol{J}_j(\boldsymbol{q})$ is the robot's translational Jacobian for point $j$ on the robot at configuration $\boldsymbol{q}$, calculated by $\boldsymbol{J}_j(\boldsymbol{q}) = \frac{\partial \boldsymbol{FK}_j(\boldsymbol{q})}{\partial \boldsymbol{q}}$.

The $t(\boldsymbol{q_p}, \boldsymbol{q_c})$ is the nominal duration from (7). Then (7) can be updated with the speed bound ($v_{max}$) allowed by SSM:

$$t(\boldsymbol{q_p}, \boldsymbol{q_c}) = \sum_{q=q_p}^{q_c} \max_{i,j} \frac{v_{robot}(q,i,j,t_n)}{v_{max}(q,i,j,t_n)} \frac{\|dq\|}{\|q_c - q_p\|} t(\boldsymbol{q_p}, \boldsymbol{q_c}), \quad (13)$$

where the range between $\boldsymbol{q_p}$ and $\boldsymbol{q_c}$ is divided into configurations of spacing $\boldsymbol{dq}$ and the time to complete each $\boldsymbol{dq}$ is summed. The $t_n$ is the nominal time to reach $\boldsymbol{q}$:

$$t_n = t_p + \frac{\|q - q_p\|}{\|q_c - q_p\|} t(\boldsymbol{q_p}, \boldsymbol{q_c}). \quad (14)$$

If $v_{robot}/v_{max}$ becomes greater than a user selected threshold, then predicted human poses over a short window into the future ($\Delta t$) can be considered by (13) to see if the delay due to human proximity is predicted to be temporary:

$$\frac{v_{robot}(q,i,j,t_n)}{v_{max}(q,i,j,t_n)} = \min_{t_s \in [t_n, t_n+\Delta t]} \frac{v_{robot}(q,i,j,t_s)}{v_{max}(q,i,j,t_s)}. \quad (15)$$

The updated $v_{robot}/v_{max}$ from (15) can be used for $v_{robot}/v_{max}$ in (13) for the configuration in question. This allows STAP to estimate delays induced by humans and the effect of SSM. A safety controller external to STAP must enforce real-time compliance with the SSM rules.

*D. Spatio-Temporal Path Planning*

The STAP method includes a variation of RRT* to use the human-avoidance model and time-based cost function. The RRT* planner was a starting point for the proposed variation because it uniformly randomly selects new nodes ($V_{new}$) from within the robot's configuration space and converges to the optimal path. The STAP method can be used with high DOF robots, so random selection of $V_{new}$ ensures exploration of robot configurations. The planner variation in STAP has a few significant deviations from standard RRT*. This variation stores data, such as avoidance intervals, for suboptimal connections for future consideration to reduce computation. Standard RRT* discards suboptimal parent connections.

Standard RRT* includes a mechanism to connect (rewire) from $V_{new}$ to a node near $V_{new}$ ($V_{near}$) if the new connection has less cost than the current connection to $V_{near}$ [28]. This ensures trajectories converge to the optimal trajectory as nodes and connections are added to $\mathcal{G}$. In STAP, $V_{near}$ is rewired to $V_{new}$ if the new connection results in a smaller $t_{arr}$ for $V_{new}$. When this occurs, STAP also considers if connections for which $V_{near}$ is the parent can be improved, up to some number of levels (denoted $N_c$) down the solution tree in the direction of children. Additionally, when a new node is added to the solution tree, all $V_{near}$ in the neighborhood of $V_{new}$ are also checked to see if there are any better parent nodes for $V_{near}$ in the neighborhood. This is necessary for STAP because a rewiring event could reduce the $t_{arr}$ for a node, possibly making it a better parent for another node. This could result in a long chain of parent upgrades and too much computation. By upgrading parent connections only for nodes near $V_{new}$, STAP ensures that if configurations for $V_{new}$ are uniformly randomly selected from the configuration space, then each existing node has equal probability of upgrading its connection to a better parent that would reduce its arrival time. Once a connection to the goal node is found, the optimal sequence of connections from the start node to the goal node, denoted $\sigma^*$, is found to be the sequence that minimizes the goal node arrival time.

*E. Time Parameterization with Avoidance Intervals*

Once STAP determines the sequence of connections $\sigma^*$, then a time parameterization can also be determined using the connection parent times computed in (8) and (9). The time to reach the goal node would be $t_c$ for the connection whose child is the goal node. Then, looking backwards through $\sigma^*$ starting from the connection to the goal node, the time the robot should reach each connection's parent node ($t_p$) has been computed in (9). The robot velocity between consecutive connections for the $i^{th}$ robot configuration variable was limited according to:

$$|\dot{q}[i]| \leq \left|\frac{q_c[i] - q_p[i]}{t_c - t_p}\right| \quad \forall i \in [1, n]. \quad (16)$$

The connection velocity limit reduces the robot's velocity if the next connection is delayed to avoid an anticipated human pose. This prevents the robot from stopping close to a human while waiting for an avoidance interval to pass. The STAP time parameterization also allows $t_p$ for the connection whose parent node is the start node to be greater than zero. This occurs when the robot should wait at the start node before beginning motion to avoid a human within the first connection.

Fig. 3 depicts a robot path solution found using the above method. Time is shown on the horizontal axis with motion start time at the left and configurations shown on the vertical axis with initial configuration at the top and goal at the bottom. Red (shaded) blocks indicate avoidance intervals where connections between configurations are obstructed. The solid black line and green circles indicate the trajectory through the time/configuration space and waypoints. It shows the connection from $\boldsymbol{q_1}$ to $\boldsymbol{q_2}$ has reduced velocity to prevent stopping at $\boldsymbol{q_2}$ until time $t_2$. The figure also depicts the last pass time ($t_{lp}$) for the $\boldsymbol{q_1} \rightarrow \boldsymbol{q_2}$ and $\boldsymbol{q_2} \rightarrow \boldsymbol{q_3}$ connections.

IV. EXPERIMENTS

Experiments with simulated and real workcells, each having a robot manipulator and a human worker, were used to validate the STAP method presented in section III. One workcell used for testing included a Comau e.Do 6DOF-6R robot sitting on a table, shown in fig. 4A. Live tests were also performed with this cell equipped with two depth cameras, outlined by blue ovals in fig. 4A, to sense the human's real-time location. While the e.Do robot has full rotation about its based, the robot is restricted to move in the half of the cell nearest the human to ensure robot/human interaction. The second workcell was equipped with a UR10e 6DOF-6R robot mounted hanging over a table, as shown in fig. 4B. Use of these two workcells allowed exploration of impacts on STAP planning by: 1) different robot maximum joint velocity limits and 2) different robot configurations (e.g., base mount, tasks / robot paths / poses). The UR10e has a maximum joint velocity about 4 times greater than the e.Do for any joint. A computer

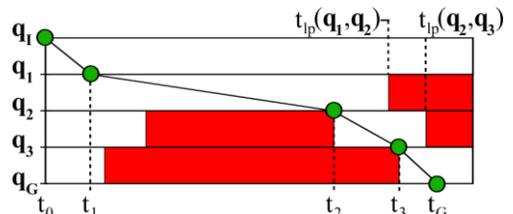

Figure 3. A sequence of nodes/connections from an initial configuration to a goal configuration indicating avoidance intervals for connections.

with an Intel i9 processor used up to 16 CPU cores for path planning.

The robot and human performed three experimental tasks together, denoted task A, B, and C. In each task, the human reached for targets, shown as red rectangles in fig. 4 and fig. 5. In fig. 5, human wrist and robot end-effector traces are solid blue and dashed green arrows, respectively. In task A, the human retrieved an item from a shelf, as shown in fig. 5A, while the robot tried to move an item from the right side of the workcell to the left side, from the human's point of view. In task B, the human retrieved an item from the right side of the table, brought it back to table in front of them, and then returned it. Meanwhile, the robot moved an item from the left to right side of the cell. The goal pose of the robot's gripper is between the human's forearm and table, as shown in fig. 5B. Task C has the human extend their arms in front, pointed at the robot, with palms together and then sweep the right arm up and left arm down 45°, as shown in fig. 5C, over 5 seconds, then back together over 5 seconds, repeating twice. The human creates a narrow passageway of time-varying size through the configuration space. Meanwhile, the robot moves an item from left to right across the cell. Tasks A and B represent more realistic motions in an industrial HRC workcell, while C presents greater challenge. In all tasks, human motion prevents the robot from reaching the goal without delay. Each task was performed 10 times in the simulated and live e.Do cells, and simulated UR10e cell. The live tests used real-time human motions sensed with the depth cameras using the method in [29] while the simulations used pre-recorded human motion to emulate a real human. Simulations provide results under ideal conditions, not including possible variations in the human's behavior caused by the interaction with the moving robot. Live tests show how the planning methods accommodate such variations in real human motions.

The STAP method was tested with waypoint timing assigned either as: 1) the output times directly from STAP time parameterization in (8) and (9) (denoted "STAP-PT" in results), or 2) the timing generated by applying Iterative Parabolic Time Parameterization (IPTP) to the STAP generated waypoints (denoted "STAP-IPTP" in results) [30]. The IPTP assigns timing so the robot will reach each waypoint as quickly as possible. These two methods only used the "offline proactive planning" block and not the "online reactive planning" block in fig. 1. For a baseline, tests were performed using the robot trajectory generated by three existing planners. The STOMP planner was used like in [15] to deform the direct trajectory, attempting to avoid predicted human motion with repulsion inversely proportional to robot-human separation distance. The Bi-directional Transition-based Rapidly-Exploring Random Trees (BiTRRT) planner repeated planning as the robot approached the real-time human location, making the robot react to the human's real-time pose [31]. The third planner (denoted "HOVs" in results) is a variant of RRT that tried to avoid all 3D points the human will occupy during the task [6]. The planning time allowed for each planner was 60 seconds to allow convergence near to their unique optimal trajectories.

In all tests, a real-time SSM safety controller applied robot speed reduction according to (11)-(13). The SSM safety controller reduced robot speed to a fraction of planned speed:

$$\dot{q} = \min_{i,j} \frac{v_{max}(q,i,j)}{v_{robot}(q,i,j)} \frac{q_c - q_p}{t_c - t_p}, \quad (17)$$

where $q$ is the robot's current configuration, in joint space, $q_p$ and $q_c$ are the configurations at the start and end of a planned connection, respectively, and $v_{max}(q,i,j)$ and $v_{robot}(q,i,j)$ use the human's current pose instead of prediction in (11)-(13). The $\dot{q}$ is the robot velocity allowed by the SSM safety mode. The ratio $v_{max}/v_{robot}$ in (17) reduces speed while $v_{robot}/v_{max}$ in (13) increases time duration. The SSM parameters were $T_r$, $a_s$, and $\underline{D}$ of 0.15s, 0.1m/s$^2$, and 0.2m, respectively. For each test, we measured the task completion time (in seconds) and the average human-robot relative distance during the trajectory execution (in meters), aiming to show that STAP can reduce the robot cycle time by avoiding severe slowdowns caused by robot proximity.

## V. RESULTS AND DISCUSSION

The results of experiments with the e.Do and the UR10e are shown in fig. 6. Fig. 6A shows the average duration the robot took to perform each trajectory. These results show that using the STAP method with IPTP resulted in trajectories that took 47% less time on average compared to the other methods, with the exception of Task A – e.Do simulated. The average time reduction by using STAP versus others was 44% for simulated e.Do, 45% for live e.Do, and 52% for simulated UR10e tests. The faster UR10e yields better time reduction results compared to the slower e.Do robot. Fig. 6B shows average robot-human separation distance during tests. Since prior works consider robot/human separation distance an important metric, it is assumed that the average separation distance metric is directly related to the human's level of comfort with robot motion [8,9-17,27]. The results show that the STAP method with either timing variation ("STAP-IPTP" or "STAP-PT") resulted in higher average robot-human separation distance. This result means the robot spent less time near the human in all tests except Task C in the UR10e Sim.

The estimated robot trajectory durations output from the STAP method were also much closer to the actual durations observed in all tests. The trajectory generated by the STAP method with IPTP took only 13% more time than the estimate, on average. The STAP method using the time parameterization of (8) and (9) had 73% average duration estimation error. The STOMP and BiTRRT planners generated estimation errors of 646% and 876%. Their estimation errors are large because they planned to do the trajectory as fast as possible but had to react to the human. The HOVs planner estimated infinite path time because its robot paths could not avoid all anticipated human occupancy volumes. A significance of these results is STAP's potential impact on task scheduling. I.e., the robot duration estimates can be used to schedule robot tasks to better match the sequence of other machines or humans in an HRC workcell. Trajectory duration estimates are also useful in robot task planning where the robot can select a next action that minimizes anticipated delay caused by humans. Planner performance was also collected for the STAP method on Task C, allotting 500 planner iterations to generate the trajectory. Fig. 7 shows evolution of path cost, or estimated trajectory duration, averaged over generation of 100 plans. It indicates the path after 180 iterations was within 15% of optimal.

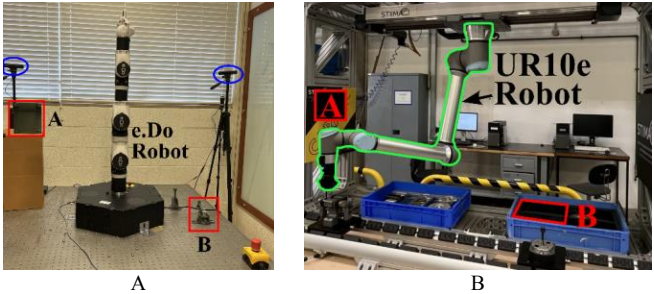

Figure 4. (A) shows the e.Do workcell and (B) shows the UR10e workcell.

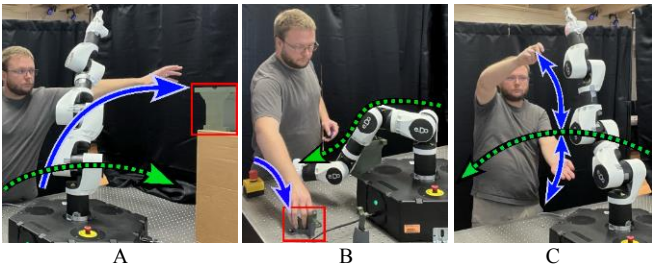

Figure 5. Experimental robot/human tasks.

A weakness of the STAP method is computation time. The first 180 iteration of STAP applied to Task C took about 30 seconds. STAP performs $N_c n^2$ times more computations per iteration than RRT* due to the STAP planner variation, where $n$ is proportional to the number of nodes at a given iteration and $N_c$ was defined in section III.D. Additional computation also stems from processing a predicted human model that is more complex than prior works. The predicted human model has many data points due to consideration of time-varying occupancy. Future works will investigate functional representations of the human model that can estimate future robot/human intersection with less computation. STAP's computation time limits its use to tasks in which human motion is predictable, e.g., cyclic manufacturing tasks. In these scenarios, STAP can plan offline based on the task definition and human task training data. During execution, if the human motion differs too much from the prediction, then the robot could use reactive planning methods, such as BiTRRT, to replan its path.

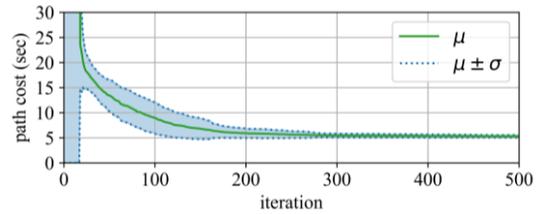

Figure 7. STAP – Task C: Evolution of average planner cost (trajectory duration estimate) and standard deviation over 500 planning iterations.

To summarize, STAP outperformed other methods by anticipating impasses due to human occupancy and delays caused by the SSM safety controller. This resulted in less-interrupted robot paths and a larger human-robot average separation distance. In addition, consideration of impasses and SSM effects allowed STAP to estimate execution times more accurately. Therefore, STAP improves HRC by providing outputs that mitigate production delay and human discomfort in predictable tasks.

## VI. CONCLUSION

The proposed method combines predicted human motion sequences and proactive robot path planning. It includes a spatio-temporal occupancy map to represent anticipated human poses and a time-avoidance cost function and variation on RRT*. The goal of the STAP method is to mitigate production delays and reduce human discomfort in an HRC workcell. Results showed the STAP method generates trajectories of shorter duration in an HRC setting. Additionally, trajectories generated by STAP result in the robot spending less time close to the human. The STAP method also outputs an estimate of the robot trajectory duration, which is useful in arriving at proactive-n-reactive robot sequencing. In future work, more compact forms of the spatio-temporal human occupancy map will be explored to address the computational challenges of STAP and a broader experimental validation will assess the efficacy of the method across different subjects and tasks.

## ACKNOWLEDGMENT

This work was made possible with funding from the National Science Foundation and collaboration with researchers at the Istituto di Sistemi e Tecnologie Industriali Intelligenti per il Manifatturiero Avanzato, Consiglio Nazionale delle Ricerche (STIIMA-CNR Researcher Director: Irene Fassi). Any opinions, findings and conclusions expressed are those of the researchers and do not necessarily reflect the views of the National Science Foundation.

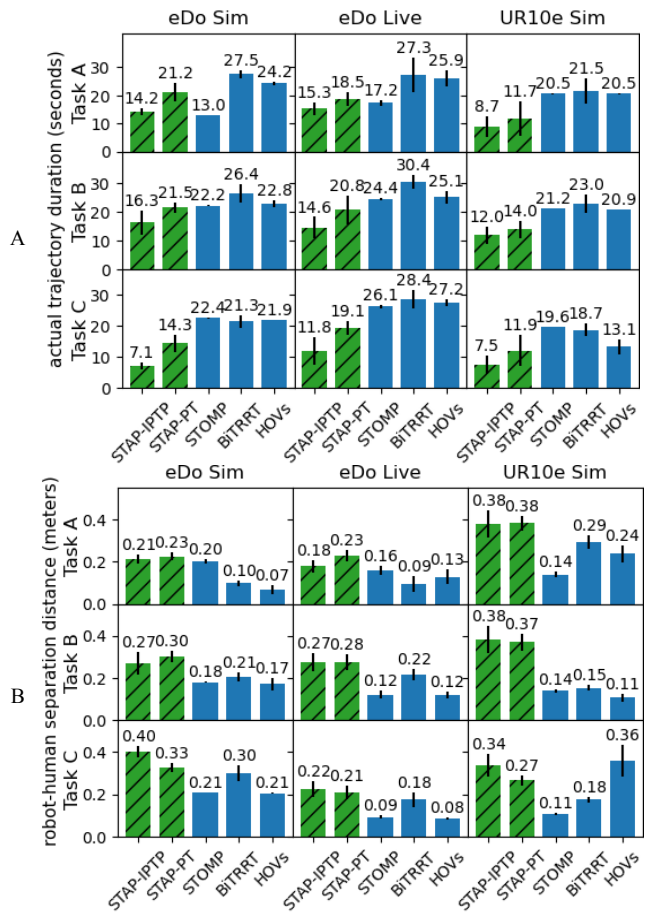

Figure 6. (A) Average trajectory durations and (B) average robot/human separation distance, with mean values indicated above the bars and standard deviations indicated by magnitude of the smaller black bars. Green hatched bars correspond to the STAP method with either timing option.